\documentclass{article}

\usepackage{arxiv}

\usepackage[utf8]{inputenc} 
\usepackage[T1]{fontenc}    
\usepackage{hyperref}       
\usepackage{url}            
\usepackage{booktabs}       
\usepackage{amsfonts}       
\usepackage{nicefrac}       
\usepackage{microtype}      
\usepackage{lipsum}
\usepackage{algorithm} 
\usepackage{algpseudocode} 
\usepackage{graphicx}
\usepackage{caption} 
\usepackage{array}
\usepackage{tabularx}
\usepackage{mathtools}
\usepackage{hhline}

\title{Channel-wise Hessian Aware trace-Weighted Quantization of Neural Networks}

\author{
  Xu Qian\\ Intel\\ xu.qian@intel.com
   \And
 Victor Li \\ Intel\\ victor.y.li@intel.com
  \And
 Crews Darren S
 \\ Intel\\ darren.s.crews@intel.com
}

\begin{document}
\maketitle

\begin{abstract}
Second-order information has proven to be very effective in determining the redundancy of neural network weights and activations. Recent paper proposes to use Hessian traces of weights and activations for mixed-precision quantization and achieves state-of-the-art results. However, prior works only focus on selecting bits for each layer while the redundancy of different channels within a layer also differ a lot. This is mainly because the complexity of determining bits for each channel is too high for original methods. Here, we introduce Channel-wise Hessian Aware trace-Weighted Quantization (CW-HAWQ). CW-HAWQ uses Hessian trace to determine the relative sensitivity order of different channels of activations and weights. What's more, CW-HAWQ proposes to use deep Reinforcement learning (DRL) Deep Deterministic Policy Gradient (DDPG)-based agent to find the optimal ratios of different quantization bits and assign bits to channels according to the Hessian trace order. The number of states in CW-HAWQ is much smaller compared with traditional AutoML based mix-precision methods since we only need to search ratios for the quantization bits. Compare CW-HAWQ with state-of-the-art shows that we can achieve better results for multiple networks.
\end{abstract}

\section{Introduction}
In recent years, real-time machine-learning applications have significantly increased. However, running deep neural networks on the edge require a lot of memory band-with to fetch the weights and computation to do dot products. For example, an inference of MobileNetV2\cite{s2018mobilenetv2} involves 6.9M weights and 585M floating point operations.

To reduce the amount of calculation, several approaches have been proposed, like pruning\cite{he2018amc}, low-rank approximation\cite{denton2014exploiting} and quantization. Among these approaches, quantization\cite{wang2018haq}\cite{lin2017accurate}\cite{courbariaux2015binaryconnect}\cite{zhou2017incremental}\cite{rastegari2016xnornet} is the most hardware friendly method. After quantization, the original activations and weights are represented with only QBN-bit (quantization bit number) fixed-point. Then neural network inferences can be performed using cheaper fixed-point multiple-accumulation operations.

In order to increase inference speed and energy efficiency on the edge, weights and activations of neural networks need to be quantized to low precision\cite{han2015deep}\cite{zhu2016trained}. Prior works use the same QBN for all layers\cite{choi2018pact}\cite{jacob2017quantization}\cite{zhou2016dorefanet}, which would introduce a huge decrease in accuracy as bit goes down.

A possible solution here is to use mixed-precision quantization, where higher precision is used for more sensitive layers of the network, and lower precision for less sensitive layers. However, the search space of choosing a QBN for each layer is too large for exhaustive search. For a widely used ResNet-50 model\cite{he2015deep}, the size of the search space is about $4^{50}$(Suppose the bit-precisions are chosen among $\{1, 2, 4, 8\}$). HAWQ-V2\cite{dong2019hawqv2} and some recent works\cite{shen2019qbert}\cite{dong2019hawq} find that second order information, like Hessian traces of different layers are great indicators of the layers’ sensitivity to quantization error. Given two layers ${L_{i}}$ and ${L_{j}}$ with average Hessian trace ${Tr_{i}}$ > ${Tr_{j}}$, we should set quantization precision ${q_{i} \geq  q_{j}}$. This method hugely decreased the complexity of the problem from $4^{50}$ to $2.3\times10^{4}$. HAWQ-V2 uses Hessian trace as criterion for layer's sensitivity and is able to outperform most model quantization methods.

However, HAWQ-V2 only focuses on selecting QBN for different layers, while the redundancy of channels within a layer also differs a lot\cite{choukroun2019lowbit}\cite{krishnamoorthi2018quantizing}\cite{banner2018posttraining}. A more fine-grained mixed precision method would be using mixed-precision for different channels. But the search space is much larger for channel-wise mixed precision problem. For ResNet-50, the search space is about $4^{27000}$. Even with HAWQ-V2, the search space is still as high as $1.3 \times 10^{15}$.This search space is even larger if we consider using mixed precision for both weights and activation channels. Traditional methods are unable to solve this kind of problem.

A recent paper AutoQ\cite{lou2019autoq}, proposes to use a hierarchical-DRL-based agent to solve this problem. AutoQ automatically search a QBN for each weight kernel and choose a QBN for each activation layer. However, the search space is too huge, so directly apply Reinforcement learning to solve channel-wise mixed precision problem would probably result in sub-optimal results. Also, the author did not consider using mixed-precision for different channels of activations.

In this paper, we introduce a Reinforcement learning based Channel-wise Hessian Aware trace-Weighted Quantization method (CW-HAWQ). We first calculate the Hessian trace of all channels and sort the channels according to the trace. Then we use DDPG\cite{lillicrap2015continuous} agent to search for the optimal ratios of different QBNs and assign QBN to channels according to the sorted trace. In CW-HAWQ, we perform channel-wise mixed precision quantization on both weights and activations, which is more fine-grained than HAWQ-V2 and AutoQ.

Also, traditional Reinforcement learning based quantization methods, like AutoQ and HAQ\cite{wang2018haq}, search for the QBNs of different layers or channels. For ResNet50, the agent states are up to 50 for layers and 27000 for channels. The complexity is relatively high. In CW-HAWQ, our states only consists of the ratios of different QBNs ($\{2, 3, 4, 5, 6, 7, 8\}$), which is much smaller than AutoQ and HAQ. Hence the agents are more likely to find a good solution during the process. 

\section{Related Works}
\textbf{Low-bit Quantization.}
In the past few years, a lot of researchers focus on increasing the accuracy for low-bit quantization on weights and activations. In PACT\cite{choi2018pact}, the author proposes to use a trainable clipping threshold to clip the activations before quantization. Using PACT, both weights and activations can be quantized to 4-bits of precision while still achieve accuracy comparable to full precision networks across a range of popular models and datasets.

However, the accuracy of PACT decreased a lot for lower bit weight quantization, especially for 2-bit. Direct 2-bit uniform quantization on weights would introduce asymmetric problem, so in Ternary network quantization\cite{zhu2016trained}, the author uses 3 bins (-1, 0, 1) to quantize the weights symmetrically. Also, ternary network uses a threshold function to clip the weights before quantization and achieves decent result on a variety of networks. Statistics-aware weight binning (SAWB)\cite{choi2018bridging}, on the other hand, fully utilizes the 4 bins of 2-bit quantization and achieves better results than ternary network. SAWB finds the optimal scaling factor that minimizes the quantization error based on the statistical characteristics of weight distribution without performing an exhaustive search.\\
\\
\textbf{Second order information based Quantization.}
Recently a lot of works focused on layer-wise mixed-precision quantization. In HAWQ\cite{dong2019hawq}, the author proposes to use second order Eigen value as indicator of the layer’s sensitivity to quantization error. The main idea is to assign higher QBN to layers that are more sensitive, and lower QBN to less sensitive layers. This can significantly reduce the exponential search space for mixed-precision quantization, since a layer with higher Hessian eigenvalues cannot be assigned lower bits, as compared to another layer with smaller Hessian eigenvalues. However, HAWQ only uses the top Hessian eigenvalue as a measure of sensitivity, and it ignores the rest of the Hessian spectrum. So in HAWQ-V2, the author uses Hessian trace instead of eigenvalues and achieves better results.\\
\\
\textbf{AutoML based Quantization.}
Recent works\cite{baker2016designing}\cite{zoph2017learning} take advantage of DRL to automatically architect CNNs for higher inference accuracy. Their network architectures outperform many human-designed neural networks. In HAQ, author proposes to use DRL DDPG agent to search for the optimal bits for each layer. Based on this, a most recent paper AutoQ proposes a two-level hierarchical DRL technique to automatically quantize the weights in the channel-wise manner and the activations in the layer-wise manner. AutoQ automatically quantize each weight kernel and each activation layer of a pre-trained CNN model for mobile devices by hierarchical DRL. AutoQ is able to outperform HAQ for multiple networks.

\section{Method}
Hessian trace is a great indicator of a neural network block’s sensitivity to quantization error\cite{dong2019hawqv2}. For channel-wise mixed-precision quantization, using Hessian trace can significantly reduce the search space, since a layer with higher Hessian trace cannot be assigned lower bits, as compared to another layer with smaller Hessian trace. With the help of Hessian trace, we are able to sort the channels and assign bits to these channels accordingly. However, the search space is still huge after Hessian trace analysis because Hessian trace does not give the exact bit precision of a given layer. 

We model this task as a Reinforcement learning problem. We use the actor-critic model with DDPG agent to give the action: ratios of different QBNs. With the prior knowledge of Hessian trace, the states of the agent only contain the ratios of different bits, in our experiments, ratios for \{2, 3, 4, 5, 6, 7, 8\}. This is much smaller compared with traditional AutoML based quantization methods where each layer or channel has an individual state like HAQ and AutoQ. 

This section is organized as follows: first we will introduce the method to generate the trace of different channels and analysis the traces of ResNet18 as an example. Then we will discuss in details about the DRL method to determine the ratios for QBNs.

\subsection{Channel-wise trace weighted quantization}
Second order information has proven to be very effective in determining layer’s sensitivity to quantization error. In HAWQ-V2, the author proposes to use the average Hessian trace to determine the relative sensitivity order of different layers and achieves state-of-the-art results. 

However, computing the Hessian trace may seem a prohibitive task, as we do not have direct access to the elements of the Hessian matrix. Furthermore, forming the Hessian matrix explicitly is not computationally feasible\cite{dong2019hawqv2}. So in HAWQ-V2, the author uses Hutchinson algorithm\cite{article} to estimate the Hessian trace of a neural network weight layer as the following equation shows:
\begin{equation}
Tr(H_w)\approx\frac{1}{m}\sum_{i=1}^{m}{z_{i}}^{T}H_{w}z^{i}=Tr_{Est}(H_{w})
\tag{1}\label{eq:1}
\end{equation}
where $z_{i}$ stands for random vector and $H_w$ stands for the weight Hessian trace of a given layer. $m$ is the total number of iterations to get a decent estimation of the Hessian trace. In HAWQ-V2, this algorithm shows good convergence properties.

Based on this insight, we modify the equation to calculate the trace for different weight channels:
\begin{equation}
\begin{aligned}
Tr(H_{w}^{j})\approx\frac{1}{m}\sum_{i=1}^{m}{{z_{i}^{j}}}^{T}H_{w}{z_{i}^{j}}=Tr_{Est}(H_w^{j})\\
{z_{i}^{j}} = Mask_{j^{th}}*z_{i}, j \in[0, output\ channels]
\end{aligned}
\tag{2}\label{eq:2}
\end{equation}
where ${z_{i}^{j}}$ is the masked random vector for the $j^{th}$ channel of a given layer. $H_{w}^{j}$ is the weight Hessian trace of $j^{th}$ channel. The $j^{th}$ channel of $Mask_{j^{th}}$ is set to 1 while the others set to 0.

Similarly for activations, the average trace for the $j^{th}$ layer can be calculated with:
\begin{equation}
Tr(H_{a})\approx\frac{1}{N}\sum_{i=1}^{N}{z_{i}}^{T}H_{a}(x_{i})z_{i}=Tr_{Est_{j}}(H_a)
\tag{3}\label{eq:3}
\end{equation}
where $x_i$ is the $i_{th}$ batch of activations. $z_{i}$ is the corresponding components of $z$ w.r.t. $x_i$. $H_a$ stands for the activation Hessian trace of a given layer. $N$ is the number of batches and random vectors to get a decent estimation of activation Hessian trace.

Then we can get the average activation trace of the $j^{th}$ channel using:
\begin{equation}
\begin{aligned}
Tr(H_{a}^{j})\approx\frac{1}{N}\sum_{i=1}^{N}{z_{i}^{j}}^{T}H_{a}(x_{i})z_{i}^{j}\\
{z_{i}^{j}} = Mask_{j^{th}}*z_{i}, j \in[0, output\ channels]
\end{aligned}
\tag{4}\label{eq:4}
\end{equation}
where $H_{a}^{j}$ is the activation Hessian trace of $j^{th}$ channel.

\begin{figure}[ht]
\centering
\begin{tabular}{ccc}
\includegraphics[width=6cm]{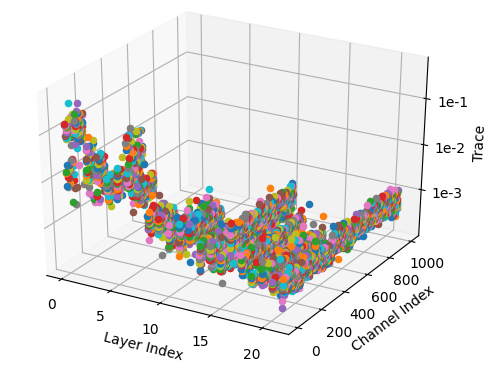}&
\includegraphics[width=6cm]{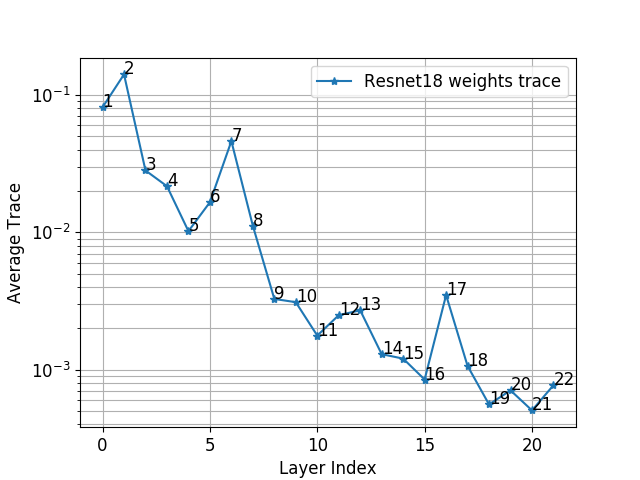}\\
(a)&(b)\\
\includegraphics[width=6cm]{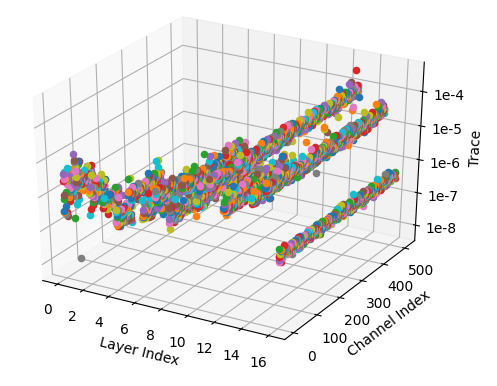}&
\includegraphics[width=6cm]{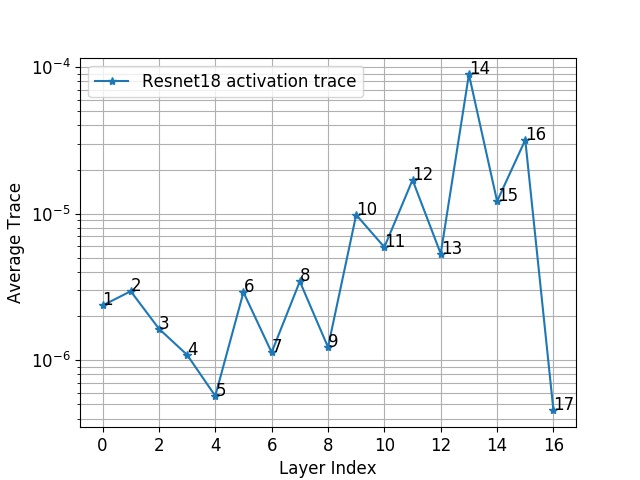}\\
(c)&(d)
\end{tabular}
\caption{Average Hessian trace of different blocks in ResNet18 on ImageNet (a) Average trace of different weight channels; (b) Average trace of different weight layers; (c) Average trace of different activation channels; (d)Average trace of different activation layers}
\label{fig:0}
\end{figure}

\begin{figure}[ht]
\centering
\begin{tabular}{ccc}
\includegraphics[width=6cm]{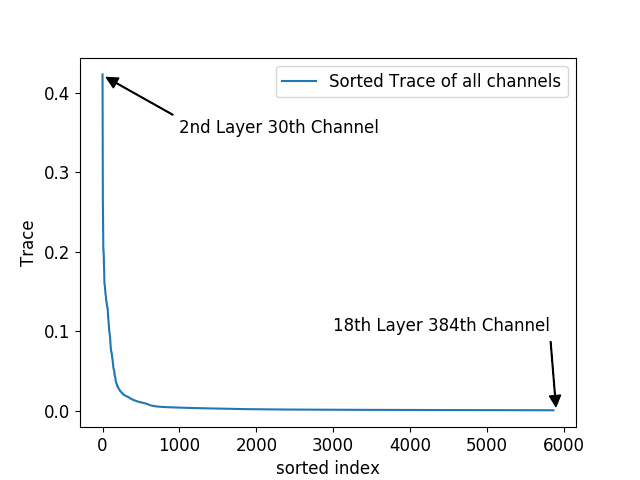}&
\includegraphics[width=6cm]{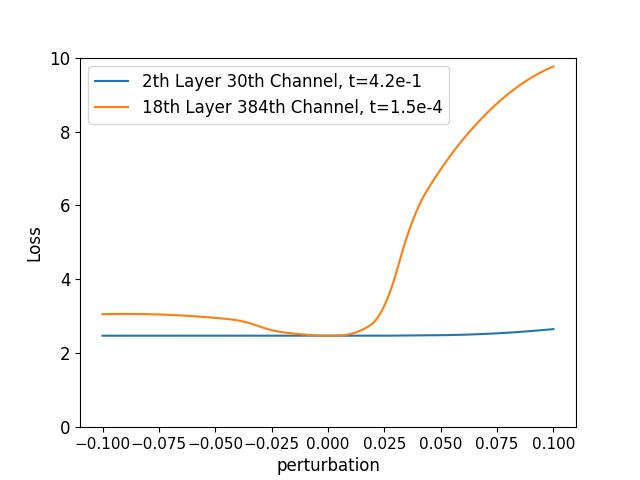}\\
(a)&(b)\\
\end{tabular}
\caption{Hessian trace analysis (a) Sorted average Hessian trace in descending order; (b) The loss landscape of channels with min and max trace}
\label{fig:1}
\end{figure}

We have incorporated the above approach and computed the average Hessian trace for different layers and channels of ResNet18, as shown in Figure \ref{fig:0}. As one can see, there is a significant difference between average Hessian traces for different layers. The traces of different channels within a layer also vary a lot. This conclusion works both for weights and activations.

We sorted the weight Hessian trace of ResNet18 in descending order as Figure \ref{fig:1}(a) shows. The difference of different channels can be as large as $10^3$.

To better illustrate this, we have also plotted the loss landscape of ResNet18 by adding perturbation to the pre-trained model as Figure \ref{fig:1}(b). It is clear that different layers have significantly different “sharpness.”  For the $30^{th}$ channel of the second layer, the average trace is about $4.2\times 10^{-1}$. The loss changes significantly as we add perturbation to this channel, so we need to use higher bits for this channel. And for the $384^{th}$ channel of the $18^{th}$ layer, the average trace is about $1.5\times 10^{-4}$. The loss landscape is relatively flat so we can quantize this channel more aggressively.

\begin{figure}[ht]
  \centering
  \includegraphics[width=12cm]{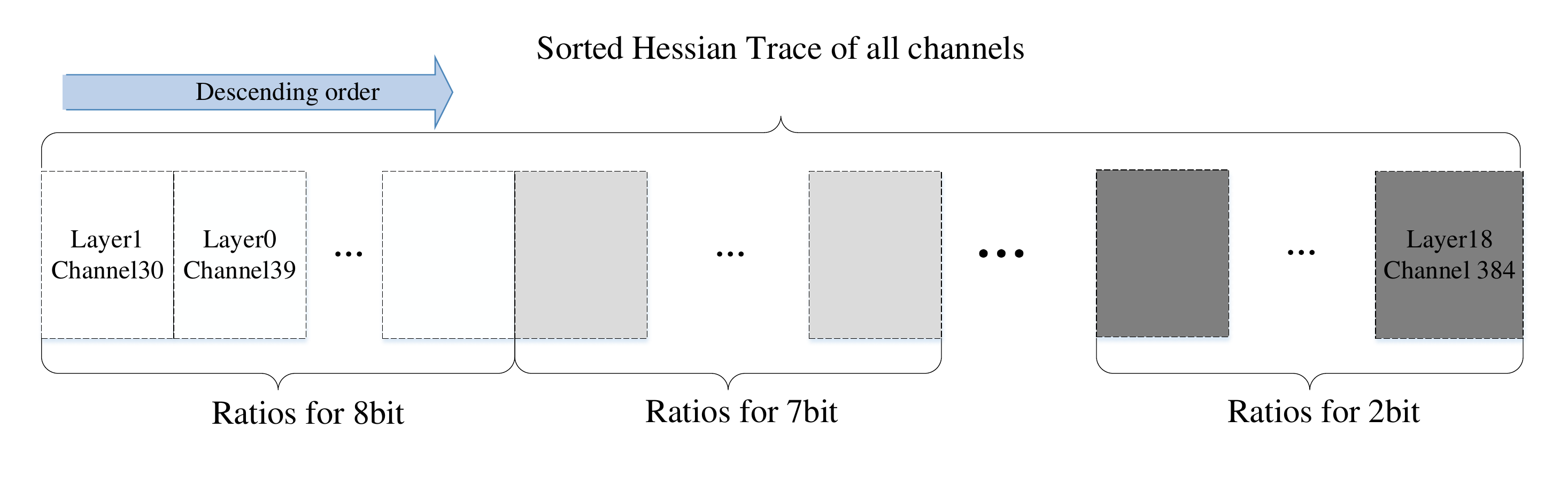}\\
  \caption{Sorted Hessian trace of all channels}
  \label{fig:2}
\end{figure}

\begin{figure}[ht]
  \centering
  \includegraphics[width=12cm]{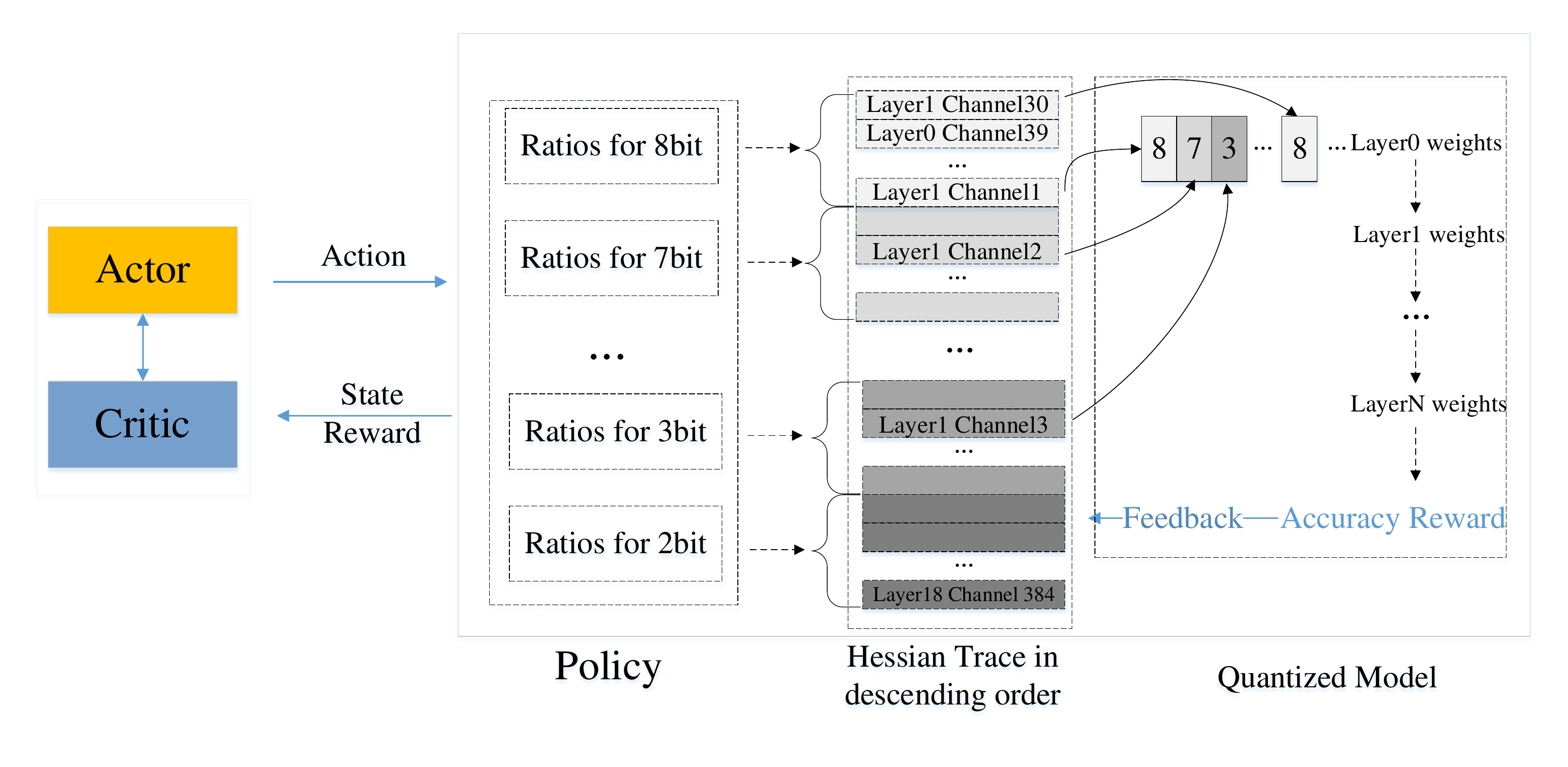}\\
  \caption{An overview of our framework. We leverage the Reinforcement learning to automatically search the ratios for different bits and allocate the bits to different channels according to the sorted Hessian trace}
  \label{fig:3}
\end{figure}
\subsection{Quantization bit allocation}
However, a major drawback for Hessian trace analysis is that it does not provide the specific bit precision setting for different channels. For example, in Figure \ref{fig:0}, we can clearly see that some channels have significant higher average trace than the others. Although it is obvious that we should allocate higher QBNs for these channels to increase accuracy, we still cannot get a specific bit precision setting.

After using Hessian trace to decide the relative order of different channels. We can sort the channels according to their Hessian traces as Figure \ref{fig:2} depicts.

Clearly, we should set higher bits to the channels with higher traces. The only parameters missing are the ratios of different QBNs. In this paper, we model this task as a DRL problem as  as Figure \ref{fig:3} shows. We use the actor-critic model with DDPG agent to give the action: ratios for different QBNs. Then we allocate the bits to the channels according to the sorted Hessian trace. In our experiments, we use compression ratio as constraint and accuracy as target to search for the optimal QBN ratios. Below describes the details of our implementation.

\subsubsection{DDPG agent}
In the previous section, we have calculate the Hessian trace of different channels. Leveraging the sorted Hessian trace list as Figure \ref{fig:2} shows, we iteratively search the ratios for $\{2, 3, 4, 5, 6, 7, 8\}$.\\
\\
\textbf{Observation (State Space).} In this process, there is no need for extra information from the network because it is implied in the sorted Hessian trace list. So the state space is relatively simple, in this paper we introduce a 5 dimension feature vector $O_{k}$ as our observation:\\
\centerline{$[k, n_{remains}, s_{i}, e_{i}, a_{k-1}]$}\\
where $k$ is the channel index, $s_{i}$ and $e_{i}$ are the starting index and ending index of the last bit in the sorted Hessian trace list, $n_{remains}$ is the number of remaining parameters, and $a_{k-1}$ is the action from the last step.
For each dimension in the observation vector $O_{k}$, we normalize it into $[0, 1]$ to make them in the same scale.\\
\\
\textbf{Action Space.} We use a continuous search space to determine the ratios for different QBNs. At the $k^{th}$ time step, we take action to search for the $k^{th}$ bit ratio.

Due to the fact that all ratios need to sum up to exact $100\%$, the Markov process in this paper is as follows: we first search the ratios of 2-bit and quantize corresponding channels according to the sorted Hessian trace. Then search the ratios of 3-bit in the remaining channels and quantize these channels accordingly. Then we repeat this operation for 4-bit to 7-bit. Then we quantize the remaining weights to 8-bit.

We encourage our agent to meet the compression constraint by limiting the action space. Every time our RL agent generates ratio for a QBN, we measure the amount of resources that will be used by the quantized model. If current policy exceeds the model size constraint, we will decrease the action until the constraint is satisfied.\\
\\
\textbf{Reward Function.} After quantization, we retrain the quantized model for one more epoch to recover the performance. As we have already imposed the resource constraints by limiting the action space, we define our reward function to be exactly the retraining accuracy top-1.\\
\\
\textbf{Quantization and fine-tuning.} We linearly quantize the weights and activations of each layers using the quantization ratios searched by DDPG agent. Specially, for 2-bit weights quantization, we utilized the method from SAWB\cite{choi2018bridging} to avoid asymmetric problem.

For activation quantization, we use the method from PACT\cite{choi2018pact} where the activations are clipped with a trainable threshold.

We follow a two step quantization method. In the first step, we use DDPG agent to search for the optimal policy for activation quantization and quantize the activations accordingly. In the second step, we use the fine-tuned model from step1 to search for the optimal policy for weight quantization. Then we use the best policy to add weights quantization to the fine-tuned model from step1.\\

\textbf{Implementation Details.}
The agent consists of an actor network and a critic network. Both share the same architecture, i.e., two hidden layers, each of which has 300 units. For the actor network, we add an additional sigmoid function producing an output in the range of [0, 1]. We use a fixed learning rate of $10^{-4}$ for the actor network and $10^{-3}$ for the critic network. We train the networks with the batch size of 64 and the replay buffer size of 600. We first explores 100 episodes with a constant noise 0.5 and then exploits 700 episodes with exponentially decayed noise.

\section{Experiments}
To evaluate CW-HAWQ, we selected several CNN models including ResNet-18, ResNet-50, MobileNetV2. The CNN models are trained on ImageNet including 1.26M training images and tested on 50K test images spanning 1K categories of objects.
\begin{figure}[hbt!]
    \centering
    \includegraphics[width=7cm]{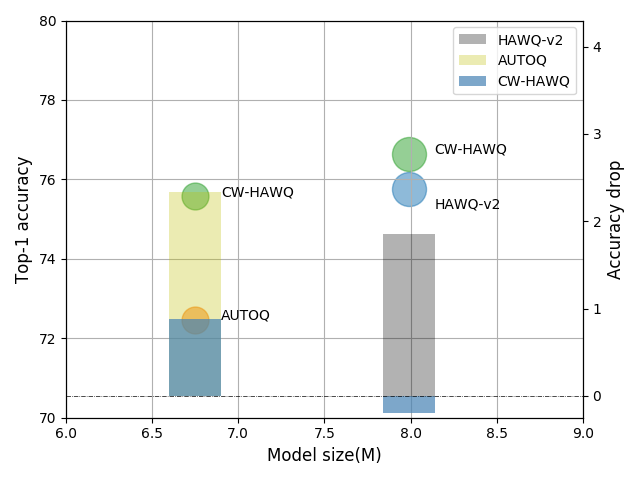}
\caption{CW-HAWQ vs AutoQ and HAWQ-V2 on ResNet50}
\label{fig:4}
\end{figure}

\begin{figure}[ht]
\centering
\begin{tabular}{ccc}
\includegraphics[width=6cm]{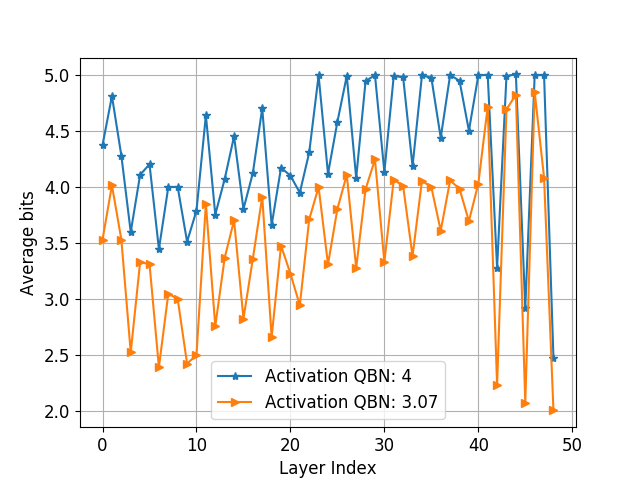}&
\includegraphics[width=6cm]{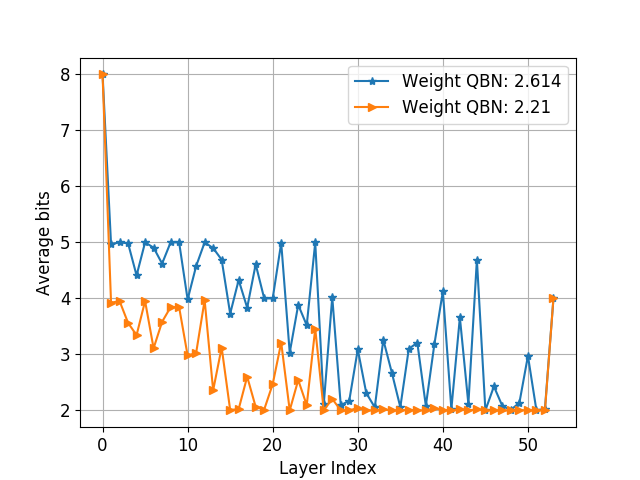}\\
(a)&(b)\\
\end{tabular}
\caption{Average QBN of different layers for ResNet50 (a) Activation QBN (b) Weight QBN}
\label{fig:5}
\end{figure}

As shown in Figure \ref{fig:4}, we apply CW-HAWQ on ResNet50 and compare the accuracy to state-of-the-art mixed precision methods (AutoQ and HAWQ-V2). It is clear that CW-HAWQ is able to outperform both method at the same compression ratio. One thing to be noted is that these three methods in the graph use different baseline, so it might not be fair to compare the accuracy top-1 directly. So I also plotted the accuracy drop bar in Figure \ref{fig:4}. As you can see, the accuracy drop is much less for CW-HAWQ. The experiment details can be found in Table \ref{table:0}. From the table, it is clear that CW-HAWQ achieved state of the art results compared with multiple methods. At 2.61 average weight QBN and 4 average activation QBN, CW-HAWQ is even able to increase the baseline accuracy by $0.2\%$.

\begin{table}[hbt!]
\caption{Results of ResNet50 on ImageNet. We abbreviate average QBN used for weights as “w-bits,” activations as “a-bits,” top-1 accuracy as “Top-1,” and weight compression ratio as “W-Comp.” Here “MP” refers to mixed-precision quantization. "Top-1 drop," column is added for a fair comparison because different methods have different baseline}
\centering
\begin{tabular}{lllllll}
\hline
Method     & w-bits & a-bits & Top-1 & Top-1 drop\footnote[2]{Top-1 accuracy drop is calculated using the baseline of each method from the original paper}
 & W-Comp & Size(MB) \\ \hline
Baseline\footnote[1]{Baseline accuracy is only for our implementation} & NA     & NA     & 76.45 & NA         & 1.00x  & 97.8     \\
PACT       &3.32    & 3.38   & 75.30 & 1.60       & 9.64x  & 10.15    \\
HAQ        &3.03    & $MP$\footnote[3]{QBN not specified in original paper is labeled as MP}
   & 75.30 & 0.85       & 10.57x & 9.22     \\
HAWQ       &2.60    &$4_{MP}$  & 75.48 & 1.91       & 12.28x & 7.96     \\
HAWQ-V2    &2.61    &$4_{MP}$  & 75.76 & 1.63       & 12.24x & 7.99     \\
AutoQ      &2.21    &3.07    & 72.47 & 2.33       & 14.48x & 6.75     \\ \hline
CW-HAWQ    &2.61    & 4      & 76.65 & -0.2       & 12.24x & 7.99     \\
CW-HAWQ    &2.21    &3.07    & 75.57 & 0.88       & 14.48x & 6.75     \\ \hline
\end{tabular}
\label{table:0}
\end{table}
\begin{table}[hbt!]
\caption{Quantization results of ResNet18 on ImageNet.}
\centering
\begin{tabular}{lllllll}
\hline
Method     & w-bits & a-bits & Top-1 & Top-1 drop & W-Comp & Size(MB) \\ \hline
Baseline & NA     & NA     & 70.47 & NA         & 1.00x  & 44.65    \\ 
PACT       & 3.18   & 4.43   & 68.1  & 2.3        & 10.06x & 4.44     \\
Uniform    & 4      & 4      & 67.3  & 2.6        & 8x     & 5.58     \\
HAQ        & 3.37   & 3.65   & 67.5  & 2.4        & 9.49x  & 4.71     \\
AutoQ      & 2.19   & 3.02   & 67.4  & 2.5        & 14.61x & 3.06     \\ \hline
CW-HAWQ    & 2.19   & 3.02   & 69.02 & 1.45       & 14.61x & 3.06     \\ \hline
\end{tabular}
\label{table:1}
\end{table}
\begin{table}[hbt!]
\centering
\caption{Quantization results of MobilenetV2 on ImageNet.}
\begin{tabular}{lllllll}
\hline
Method     & w-bits & a-bits & Top-1 & Top-1 drop & W-Comp & Size(MB) \\ \hline
Baseline & NA     & NA     & 71.8  & NA         & 1.00x  & 13.51    \\
Uniform    & 4      & 4      & 68.65 & 2.45       & 8x     & 1.69     \\
HAQ        & 3.21   & 3.92   & 68.66 & 2.44       & 9.97x  & 1.36     \\
AutoQ      & 2.26   & 3.13   & 68.68 & 2.42       & 14.16x & 0.95     \\ \hline
CW-HAWQ    & 2.26   & 3.13   & 69.85 & 1.95       & 14.16x & 0.95     \\ \hline
\end{tabular}
\label{table:2}
\end{table}
We also experimented with ResNet18, as in Table \ref{table:1}, we compared CW-HAWQ with Uniform, AutoQ and HAQ methods. For Uniform method, the QBNs are the same for all layers. Our quantized model is able to achieve about $1.6\%$ higher accuracy than AutoQ at the same compression ratio. Our Top-1 drop is about $1\%$ smaller considering AutoQ used a lower baseline.

We also tried smaller networks like MobilenetV2, as in Table \ref{table:2}. CW-HAWQ is able to achieve about $1.2\%$ better accuracy than AutoQ at the same compression ratio. The Top-1 drop is also about $0.5\%$ higher.

\bibliographystyle{unsrt}  
\bibliography{template}






\end{document}